\def\year{2016}
\pdfoutput=1

\documentclass[letterpaper]{article}
\usepackage{aaai16}

\nocopyright

\usepackage{times}
\usepackage[scaled=0.86]{helvet}
\usepackage{courier}
\usepackage{xspace}
\usepackage{booktabs}
\usepackage{subfig}
\usepackage{tabularx}
\usepackage{url}
\usepackage{amsmath,amssymb}
\usepackage{hyperref}
\usepackage{natbib}
\usepackage{cleveref}

\renewcommand{\vec}[1]{\textit{\textbf{#1}}}
\newcommand{\ve}{\vec{e}}
\newcommand{\vr}{\vec{r}}
\newcommand{\vc}{\vec{c}}
\newcommand{\vx}{\vec{x}}

\newcommand{\va}{\vec{a}}
\newcommand{\vb}{\vec{b}}

\newcommand{\comp}{\circ}
\newcommand{\conv}{\ast}
\newcommand{\corr}{\star}
\newcommand{\cconv}{\ast}
\newcommand{\concat}{\oplus}
\newcommand{\ccorr}{\star}
\newcommand{\conj}[1]{\overline{#1}}
\newcommand{\invl}[1]{\widetilde{#1}}
\newcommand{\hole}{\textsc{HolE}\xspace}
\newcommand{\transe}{\textsc{TransE}\xspace}
\newcommand{\transr}{\textsc{TransR}\xspace}
\newcommand{\transh}{\textsc{TransH}\xspace}
\newcommand{\rescal}{\textsc{Rescal}\xspace}
\newcommand{\emlp}{\textsc{ER-MLP}\xspace}
\newcommand{\hadamard}{\odot}
\newcommand{\Set}[1]{\mathcal{#1}}
\newcommand{\setdef}[2]{\left\{#1\ \middle|\, #2\right\}}
\newcommand{\transp}{\top}
\newcommand{\kron}{\otimes}
\newcommand{\R}{\mathbb{R}}
\newcommand{\SE}{\Set{E}}
\newcommand{\SR}{\Set{R}}
\newcommand{\Ss}{\textsf{s}}
\newcommand{\So}{\textsf{o}}

\newcommand{\fft}{\mathcal{F}}
\newcommand{\ifft}{\mathcal{F}^{-1}}
\newcommand{\BigO}{\mathcal{O}}
\newcommand{\loss}{\mathcal{L}}
\newcommand{\lif}{\Rightarrow} 				
\DeclareMathOperator*{\argmax}{arg\,max} 		

\usepackage{tikz}
\usetikzlibrary{arrows,topaths,calc}
\usetikzlibrary{shadows,positioning}
\usetikzlibrary{shapes}
\usetikzlibrary{backgrounds}
\usetikzlibrary{decorations.pathreplacing}
\usepackage{tkz-graph}

\usepackage{microtype}

\begin{document}
%

\newcommand{\inst}[1]{\textsuperscript{\normalfont #1}}
\title{Holographic Embeddings of Knowledge Graphs}
\author{Maximilian Nickel\inst{1,2} \and Lorenzo Rosasco\inst{1,2,3} \and Tomaso Poggio\inst{1}\\
  \inst{1}Laboratory for Computational and
  Statistical Learning and Center for Brains, Minds and Machines\\Massachusetts Institute of
  Technology, Cambridge, MA\\
  \inst{2}Istituto Italiano di Tecnologia, Genova, Italy\\
  \inst{3}DIBRIS, Universita Degli Studi Di Genova, Italy
}
\maketitle
\begin{abstract}
  Learning embeddings of entities and relations is an efficient and versatile
  method to perform machine learning on relational data such as knowledge graphs.
  In this work, we propose holographic embeddings (\hole) to learn compositional
  vector space representations of entire knowledge graphs.
  The proposed method is related to holographic models of associative memory in
  that it employs circular correlation to create compositional representations.
  By using correlation as the compositional operator, \hole can capture rich
  interactions but simultaneously remains efficient to compute, easy to train, and
  scalable to very large datasets.
  Experimentally, we show that holographic embeddings are able to
  outperform state-of-the-art methods for link prediction on knowledge graphs and
  relational learning benchmark datasets.
\end{abstract}

\noindent 

\section{Introduction}
Relations are a key concept in artificial intelligence and cognitive science.
Many of the structures that humans impose on the world, such as logical
reasoning, analogies, or taxonomies, are based on entities, concepts and their
relationships. Hence, learning from and with relational knowledge
representations has long been considered an important task in artificial
intelligence (see e.g.,
\cite{getoor:srlbook07,muggleton1991inductive,gentner1983structure,Kemp:2006:LSC:1597538.1597600,Xu06infinitehidden,richardson2006markov}).
In this work we are concerned with learning from knowledge graphs (KGs), i.e.,
knowledge bases which model facts as instances of binary relations
(e.g.,~\mbox{\textit{bornIn(BarackObama, Hawaii)}}). This form of knowledge
representation can be interpreted as a multigraph, where entities
correspond to nodes, facts correspond to typed edges, and the type of an edge
indicates the kind of the relation. Modern knowledge graphs such as
YAGO~\citep{suchanek_yago:_2007}, DBpedia~\citep{auer_dbpedia:_2007}, and
Freebase~\citep{bollacker_freebase:_2008} contain billions of facts about
millions of entities and have found important applications in question
answering, structured search, and digital assistants.
Recently, vector space embeddings of knowledge graphs have received considerable
attention, as they can be used to create statistical models of entire KGs,
i.e., to predict the probability of any possible relation instance
(edge) in the graph. Such models can be used to derive new knowledge from known
facts (link prediction), to disambiguate entities (entity resolution), to
extract taxonomies, and for probabilistic question answering (see
e.g.,~\citep{nickel_three-way_2011,bordes2013translating,krompass2014querying}).
Furthermore, embeddings of KGs have been used to support machine
reading and to assess the trustworthiness of web
sites~\citep{dong2014knowledge,dong2015trust}.
%
However, existing embedding models that can capture rich interactions in
relational data are often limited in their scalability. Vice versa, models that
can be computed efficiently are often considerably less expressive.
In this work, we approach learning from KGs within the framework of compositional vector
space models. We introduce holographic embeddings (\hole) which use the circular
correlation of entity embeddings (vector representations) to create
compositional representations of binary relational data.
By using correlation as the compositional operator \hole can capture rich
interactions but simultaneously remains efficient to compute, easy to train, and
scalable to very large datasets.
As we will show experimentally, \hole is able to outperform
state-of-the-art embedding models on various benchmark
datasets for learning from KGs.
Compositional vector space models have also been considered in cognitive science
and natural language processing, e.g., to model symbolic structures, to
represent the semantic meaning of phrases, and as models for associative
memory~(see e.g.,
\cite{smolensky_tensor_1990,plate1995holographic,mitchell-lapata:2008:ACLMain,socher2012semantic}).
In this work, we do not only draw inspiration from these models, but we will
also highlight the connections of \hole to holographic models of associative
memory.

\section{Compositional Representations}
\label{sec:comp-rep}

In this section we introduce compositional vector space models
for KGs, the general learning setting, and related work. 

Let $\SE$ denote the set of all entities and $\Set{P}$ the set of all relation
types (predicates) in a domain.
A binary relation is a subset $\SR_p \subseteq \SE \times \SE$ of all pairs of
entities (i.e., those pairs which are in a relation of type $p$). Higher-arity
relations are defined analogously.
The characteristic function ${\phi_{p} : \SE \times \SE \to \{\pm 1\}}$ of a
relation $\SR_p$ indicates for each \emph{possible} pair of entities whether
they are part of $\SR_p$.
We will denote (possible) relation instances as $\SR_p(\Ss,
\So)$, where $\Ss, \So \in \SE$ denote the first and second argument of the
asymmetric relation $\SR_p$. 
We will refer to $\Ss, \So$ as \emph{subject} and \emph{object}
and to $\SR_p(\Ss, \So)$ as \emph{triples}.

Compositional vector space models provide an elegant way to learn the
characteristic functions of the relations in a knowledge graph, as they allow to
cast the learning task as a problem of supervised representation learning.
Here, we discuss models of the form 
\begin{equation}
  \label{eq:comp}
  \Pr(\phi_p(\Ss, \So) = 1 | \Theta) = \sigma(\eta_{spo}) = \sigma(\vr_p^\transp (\ve_s \comp \ve_o))
\end{equation}
where $\vr_p \in \R^{d_r}$, $\ve_i \in \R^{d_e}$ are vector representations
of relations and entities;
$\sigma(x) = 1 / (1 + \exp(-x))$ denotes the logistic function;
$\Theta = \{\ve_i\}_{i=1}^{n_e} \cup \{\vr_k\}_{k=1}^{n_r}$ denotes the set of
all embeddings;
$\circ: \R^{d_e} \times \R^{d_e} \to \R^{d_p}$ denotes the compositional
operator which creates a \emph{composite} vector representation for the pair
$(\Ss, \So)$ from the embeddings $\ve_s, \ve_o$. We will discuss possible
compositional operators below.

Let ${x_i \in \Set{P} \times \SE \times \SE}$ denote a triple, and
${y_i \in \{\pm1\}}$ denote its label.
Given a dataset ${\Set{D} = \{(x_i, y_i)\}_{i=1}^m}$ of true and false relation
instances, we then want to learn representations of entities and relations
$\Theta$ that best explain $\Set{D}$ according to \cref{eq:comp}. This can, for instance, be
done by minimizing the (regularized) logistic loss
\begin{equation}
  \label{eq:loss}
  \min_{\Theta} \sum_{i=1}^m \log(1 + \exp(-y_i \eta_i)) + \lambda \|\Theta\|_2^2 .
\end{equation}
%
For relational data, minimizing the logistic loss has the additional advantage
that it can help to find low dimensional embeddings for complex relational
patterns~\citep{bouchard2015approximate}.
However, in most cases, KGs store only true triples and non-existing
triples can be either missing of false (open-world
assumption). In this case, negative examples can be generated by heuristics
such as the local closed world assumption~\citep{dong2014knowledge}.
Alternatively, we can use a pairwise ranking loss such as 
\begin{equation}
  \label{eq:margin-loss}
  \min_{\Theta} \sum_{i \in \Set{D}_+} \sum_{j \in \Set{D}_{-}} \max(0, \gamma + \sigma(\eta_j) - \sigma(\eta_i))
\end{equation}
to rank the probability of existing triples higher than 
the probability of non-existing ones. Here,  
$\Set{D}^+, \Set{D}^-$ denote the set of existing and non-existing
triples and $\gamma > 0$ specifies the width of the margin \citep{bordes_learning_2011}.

An important property of compositional models is that the meaning and
representation of entities does \emph{not} vary with regard to their position in
the compositional representation (i.e., the $i$-th entity has the same
representation $\ve_i$ as subject and object). Since the representations of all
entities and relations are learned jointly in \cref{eq:loss,eq:margin-loss},
this property allows to propagate information between triples, to capture
global dependencies in the data, and to enable the desired relational learning
effect. 
For a review of machine learning on knowledge graphs
see also~\citet{nickel2015review}.

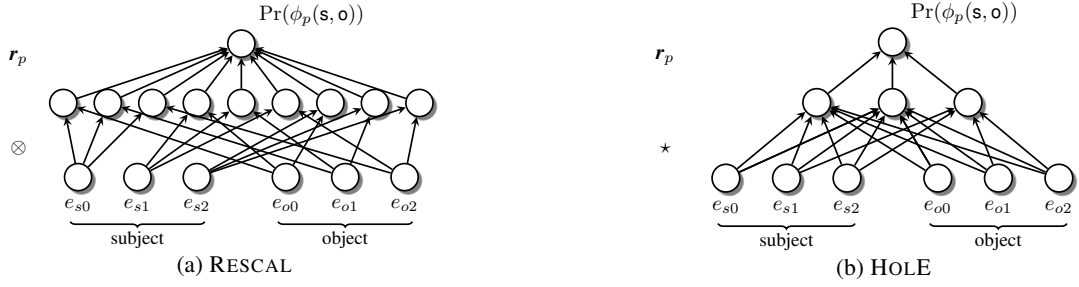
\begin{figure*}[tb]
  \centering
  \subfloat[\rescal]{
    \label{fig:nn:rescal}
    \begin{minipage}{0.33\textwidth}
      \hspace{-1em}
      \resizebox{\columnwidth}{!}{
        \begin{tikzpicture}[scale=1,baseline,thick]
          \GraphInit[vstyle=Classic]
          \tikzset{vertex/.style =
            {draw=black,shape=circle,fill=white,minimum size=13pt,circular
              drop shadow}}
          \node at (-0.5,0)[vertex,label={below:$e_{s0}$}] (s1) {};
          \node at (0.5,0)[vertex,label={below:$e_{s1}$}] (s2) {};
          \node at (1.5,0)[vertex,label={below:$e_{s2}$}] (s3) {};
          \node at (3,0)[vertex,label={below:$e_{o0}$}] (o1) {};
          \node at (4,0)[vertex,label={below:$e_{o1}$}] (o2) {};
          \node at (5,0)[vertex,label={below:$e_{o2}$}] (o3) {};
          \node at (-0.75,1.25)[vertex] (s1o1) {};
          \node at (0,1.25)[vertex] (s1o2) {};
          \node at (0.75,1.25)[vertex] (s1o3) {};
          \node at (1.5,1.25)[vertex] (s2o1) {};
          \node at (2.25,1.25)[vertex] (s2o2) {};
          \node at (3,1.25)[vertex] (s2o3) {};
          \node at (3.75,1.25)[vertex] (s3o1) {};
          \node at (4.5,1.25)[vertex] (s3o2) {};
          \node at (5.25,1.25)[vertex] (s3o3) {};
          \node at (2.25,2.25)[vertex,label=above right:{$\Pr(\phi_p(\Ss,\So))$}] (out) {};

          \node at (-1.5, 0.5) (kron) {$\kron$};
          \node at (-1.5, 2) (W) {$\protect{\vec{r}_p}$};

          \node at (-0.5,-0.75) (s1l) {};
          \node at (1.5,-0.75) (s2l) {};
          \node at (0.5,-1.05) (subl) {\footnotesize subject};
          \draw[decoration={brace,mirror},decorate] (s1l.west) -- (s2l.east);

          \node at (3,-0.75) (o1l) {};
          \node at (5,-0.75) (o2l) {};
          \node at (4,-1.05) (subl) {\footnotesize object};
          \draw[decoration={brace,mirror},decorate] (o1l.west) -- (o2l.east);

          \tikzstyle{EdgeStyle}=[->,>=stealth,thick]
          \Edge (s1)(s1o1) \Edge (s1)(s1o2) \Edge (s1)(s1o3)
          \Edge (s2)(s2o1) \Edge (s2)(s2o2) \Edge (s2)(s2o3)
          \Edge (s3)(s3o1) \Edge (s3)(s3o2) \Edge (s3)(s3o3)
          \Edge (s1o1)(out) \Edge (s1o2)(out) \Edge (s1o3)(out)
          \Edge (s2o1)(out) \Edge (s2o2)(out) \Edge (s2o3)(out)
          \Edge (s3o1)(out) \Edge (s3o2)(out) \Edge (s3o3)(out)
          \Edge (o1)(s1o1) \Edge (o1)(s2o1) \Edge (o1)(s3o1)
          \Edge (o2)(s1o2) \Edge (o2)(s2o2) \Edge (o2)(s3o2)
          \Edge (o3)(s1o3) \Edge (o3)(s2o3) \Edge (o3)(s3o3)
        \end{tikzpicture}
      }
      \vspace{-1ex}
    \end{minipage}
  }
  \hspace{7em}
  \subfloat[\hole]{
    \label{fig:nn:hole}
    \begin{minipage}{0.33\textwidth}
      \hspace{-1em}
      \resizebox{\columnwidth}{!}{
        \begin{tikzpicture}[scale=1,baseline,thick]
          \GraphInit[vstyle=Classic]
          \tikzset{vertex/.style =
            {draw=black,shape=circle,fill=white,minimum size=13pt,circular
              drop shadow}}
          \node at (-0.5,0)[vertex,label={below:$e_{s0}$}] (s1) {};
          \node at (0.5,0)[vertex,label={below:$e_{s1}$}] (s2) {};
          \node at (1.5,0)[vertex,label={below:$e_{s2}$}] (s3) {};
          \node at (3,0)[vertex,label={below:$e_{o0}$}] (o1) {};
          \node at (4,0)[vertex,label={below:$e_{o1}$}] (o2) {};
          \node at (5,0)[vertex,label={below:$e_{o2}$}] (o3) {};
          \node at (1,1.25)[vertex] (t1) {};
          \node at (2.25,1.25)[vertex] (t2) {};
          \node at (3.5,1.25)[vertex] (t3) {};
          \node at (2.25,2.25)[vertex,label=above right:{$\Pr(\phi_p(\Ss, \So))$}] (out) {};

          \node at (-1.5, 0.5) (kron) {$\ccorr$};
          \node at (-1.5, 2) (W) {$\protect{\vec{r}_p}$};

          \node at (-0.5,-0.75) (s1l) {};
          \node at (1.5,-0.75) (s2l) {};
          \node at (0.5,-1.05) (subl) {\footnotesize subject};
          \draw[decoration={brace,mirror},decorate] (s1l.west) -- (s2l.east);

          \node at (3,-0.75) (o1l) {};
          \node at (5,-0.75) (o2l) {};
          \node at (4,-1.05) (subl) {\footnotesize object};
          \draw[decoration={brace,mirror},decorate] (o1l.west) -- (o2l.east);

          \tikzstyle{EdgeStyle}=[->,>=stealth,thick]
          \Edge (s1)(t1) \Edge (s1)(t2) \Edge (s1)(t2)
          \Edge (s2)(t1) \Edge (s2)(t2) \Edge (s2)(t3)
          \Edge (s3)(t1) \Edge (s3)(t2) \Edge (s3)(t3)
          \Edge (o1)(t1) \Edge (o1)(t2) \Edge (o1)(t2)
          \Edge (o2)(t1) \Edge (o2)(t2) \Edge (o2)(t3)
          \Edge (o3)(t1) \Edge (o3)(t2) \Edge (o3)(t3)
          \Edge (t1)(out) \Edge (t2)(out) \Edge (t3)(out)
        \end{tikzpicture}
      }
      \vspace{-1ex}
    \end{minipage}
  }
  \caption{\rescal and \hole as neural networks. \rescal represents
    pairs of entities via $d^2$ components (middle layer). In contrast, \hole
    requires only $d$ components.\label{fig:nn}}
\end{figure*}

Existing models for knowledge graphs are based on the following compositional
operators:
\paragraph{Tensor Product}
Given entity embeddings $\va, \vb \in \R^d$, tensor product models
represent pairs of entities via ${\va \comp \vb = \va \kron \vb \in \R^{d^2}}$, i.e,
via all pairwise multiplicative interactions between the features of $\va$ and
$\vb$:
\begin{equation}
  \label{eq:kron}
  {[\va \kron \vb]}_{ij} = a_i b_j .
\end{equation}
Intuitively, a feature in the tuple representation $\va \kron \vb$ is ``on''
(has a high absolute magnitude), if and only if the corresponding features
of \emph{both} entities are ``on'' (See also \cref{fig:nn:rescal}). 
This allows \cref{eq:comp} to capture relational patterns such as 
\emph{liberal persons are typically members of liberal parties} since 
a single feature in $\va \kron \vb$ can encode that the subject is a 
liberal person and that the object is a liberal party.
Compositional models using the tensor product 
such as \rescal~\citep{nickel_three-way_2011}
and the Neural Tensor Network \citep{socher2013reasoning}.
have shown state-of-the-art performance for learning from KGs.
Furthermore, \citet{guu2015traversing} proposed a \textsc{Rescal}-based model to learn from paths in KGs.
\citet{smolensky_tensor_1990} introduced 
the tensor product as a way to create compositional vector representations.
While the tensor product allows to capture rich interactions, its main problem
as a compositional operator lies in the fact that it requires a large number of
parameters. Since $\va \kron \vb$ explicitly models all pairwise interactions,
$\vr_p$ in \cref{eq:comp} must be of size $d^2$. This can be problematic both in
terms of overfitting and computational demands. For instance,
\citet{nickel2014reducing} showed that linear tensor factorization can require
large $d$ to model certain relations. Since ${\vec{r}_p^\transp(\vec{e}_s \kron
  \vec{e}_o) =\ve_s^\transp R_p \ve_o}$, \citet{yang2015embedding} proposed to
use diagonal $R_p$'s to reduce the number of parameters. However, this approach
can only model symmetric relations and is not suitable to model general
knowledge graphs as $\ve_s^\transp R_p \ve_o = \ve_o^\transp R_p \ve_s$ for
diagonal $R_p$.

\paragraph{Concatenation, Projection, and Non-Linearity}
Another way to compute composite representations is via concatenation,
projection and subsequent application of a non-linear function. 
Let $\concat : \R^{d_1} \times \R^{d_2} \to \R^{d_1 + d_2}$
denote concatenation and $\psi : \R \to \R$ be a non-linear function
such as \emph{tanh}. The composite tuple representation is then
given by $\va \comp \vb = \psi(W (\va \concat \vb)) \in \R^h$, such that
\begin{equation}
  \label{eq:concat}
  [\psi(W (\va \concat \vb))]_i = \psi\left(\sum\nolimits_j w_{ij}^a a_j + \sum\nolimits_j w_{ij}^b b_j\right)
\end{equation}
where the projection matrix $W \in \R^{h \times 2d}$ is learned in
combination with the entity and relation embeddings. Intuitively, a feature in
the tuple representation $W(\va \concat \vb)$ is ``on'' if \emph{at least one}
of the corresponding features is ``on''. An advantage of this
compositional operator is that the mapping from entity embeddings to
representations of pairs is learned adaptively via the matrix $W$. 
%
%
However, the resulting composite representations
are also less rich, as they do not consider direct interactions of
features. As \citet{socher2013reasoning} noted, the non-linearity
$\psi$ provides only weak interactions while leading to a harder optimization problem.
A variant of this compositional operator which also includes a relation
embedding in the composite representation has been used in
the ER-MLP model of the Knowledge Vault~\citep{dong2014knowledge}.

\paragraph{Non-compositional Methods}
Another class of models does not (explicitly) form compositional
representations, but predicts the existence of triples from
the similarity of the vector space embeddings. In particular,
\transe~\citep{bordes2013translating} models the score of a fact as
the distance between relation-specific translations of entity embeddings:
\begin{equation}
  \operatorname{score}(\SR_p(\Ss, \So)) = -\operatorname{dist(\ve_s + \vr_p, \ve_o)} .
\end{equation}
A major appeal of \transe is that it requires very few parameters and moreover is
easy to train. However, this simplicity comes also at the cost of modeling
power. \citet{wang2014knowledge} and \citet{lin2015learning}
proposed \transh and \transr respectively, to improve the performance of \transe
on 1-to-N, N-to-1, and N-to-N relations.
Unfortunately, these models lose the
simplicity and efficiency of \transe.


\section{Holographic Embeddings}

\label{sec:hole}
In this section, we propose a novel compositional model for KGs and relational data. 
To combine the expressive power of the tensor product with the
efficiency and simplicity of \transe, we use the circular correlation 
of vectors to represent pairs of entities, i.e., we use the compositional operator:
\begin{equation}
  \label{eq:hole-comp}
  \va \comp \vb = \va \ccorr \vb ,
\end{equation}
where $\ccorr : \R^d \times \R^d \to \R^d$ denotes
\emph{circular correlation}:\footnote{For notational convenience, we use
  zero-indexed vectors.}
\begin{equation}
  [\vec{a} \ccorr \vec{b}]_k = \sum_{i=0}^{d-1} a_i b_{(k + i) \operatorname{mod} d} .
  \label{eq:ccorr}
\end{equation}
Hence, we model the probability of a triple as
\begin{equation}
  \label{eq:hole}
  \Pr(\phi_p(\Ss, \So) = 1 | \Theta) = \sigma(\vr_p^\transp (\ve_s \ccorr \ve_o)) .
\end{equation}
Due to its connections to holographic models of associative memory (which we
will discuss in the next section) we refer to \cref{eq:hole} as
\emph{holographic embeddings} (\hole) of KGs.

{
  \captionsetup{position=top}
  \begin{table}[bt]
    \caption{(a) Memory complexity and runtime complexity %
      for compositional representations with $\ve_i \in \R^d$.
      (b) Memory complexity of embedding models.}
    \centering
    \small
    \subfloat[Compositional Representations]{\label{tab:complexity}
      \noindent
      \begin{tabularx}{0.975\columnwidth}{lcXX}
        \toprule
        \textbf{Operator} & \quad $\comp$ \quad &\textbf{Memory} & \textbf{Runtime}
        \\
        & & $\vec{r}_p$ & $\vec{r}_p^\transp(\vec{e}_s \comp \vec{e}_o)$ \\
        \midrule
        Tensor Product & \quad $\kron$ \quad & $\BigO(d^2)$ & $\BigO(d^2)$ \\
        Circular Correlation & \quad $\ccorr$ \quad & $\BigO(d)$ & $\BigO(d \log d)$ \\
        \bottomrule
      \end{tabularx}
    }

    \subfloat[Embedding Models]{\label{tab:params}
      \begin{tabularx}{.975\columnwidth}{lXlr}
        \toprule 
        & & \multicolumn{2}{c}{\textbf{on FB15k}} \\
        \cmidrule(l){3-4}
        \textbf{Method} & \textbf{Memory Complexity} & $d$ & Params\\
        \midrule
        \transe & $\BigO({n_e}d + {n_r}d)$ & 100 & 1.6M\\
        \transr & $\BigO(n_ed + n_rd + n_rd^2)$ & 100 & 15.1M\\
        \emlp & $\BigO({n_e}d + {n_r}d + d_pd)$ & 200/200 & 3.3M\\
        \rescal & $\BigO({n_e}d + {n_r}d^2)$ & 150 & 32.5M\\
        \midrule
        \hole & $\BigO({n_e}d + {n_r}d)$ & 200 & 3.3M \\
        \bottomrule
      \end{tabularx}
    }
  \end{table}
}

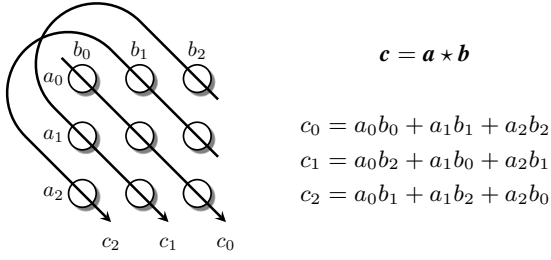
\begin{figure}
  \centering
  \begin{minipage}{0.5\columnwidth}
    \centering
    \resizebox{0.8\columnwidth}{!}{
      \begin{tikzpicture}[scale=0.95,baseline,thick,node distance=.5cm]
        \GraphInit[vstyle=Classic]
        \tikzset{vertex/.style =
          {draw=black,shape=circle,fill=white,minimum size=13pt,circular
            drop shadow}}
        \tikzset{sum/.style = {->,>=stealth,very thick}}
        \node at (0.5,1) (a3) {$a_2$};
        \node at (0.5,2) (a2) {$a_1$};
        \node at (0.5,3) (a1) {$a_0$};
        \node at (1,3.5) (b1) {$b_0$};
        \node at (2,3.5) (b2) {$b_1$};
        \node at (3,3.5) (b3) {$b_2$};
        \node at (1.5,0.5)[label=below:$c_2$] (z3) {};
        \node at (2.5,0.5)[label=below:$c_1$] (z2) {};
        \node at (3.5,0.5)[label=below:$c_0$] (z1) {};
        \node at (1,1)[vertex] (a3b1) {};
        \node at (1,2)[vertex] (a2b1) {};
        \node at (1,3)[vertex] (a1b1) {};
        \node at (2,1)[vertex] (a3b2) {};
        \node at (2,2)[vertex] (a2b2) {};
        \node at (2,3)[vertex] (a1b2) {};
        \node at (3,1)[vertex] (a3b3) {};
        \node at (3,2)[vertex] (a2b3) {};
        \node at (3,3)[vertex] (a1b3) {};

        \node at (2,4) (h1){};
        \node at (0.5,2.5) (h2){};
        \node at (1.5,3.5) (h3){};
        \node at (0,2) (h4){};

        \node at (0.5,3.5) (s1) {};
        \node at (3.5,2.5) (s2) {};
        \node at (3.5,1.5) (s3) {};

        \draw[sum] (s1) -- (z1.center);
        \draw[sum] (s2) -- (h1.center) arc(-135:45:-1.06) (h2.center) -- (z2.center);
        \draw[sum] (s3) -- (h3.center) arc(-135:45:-1.06) (h4.center) -- (z3.center);
      \end{tikzpicture}
    }
  \end{minipage}
  \begin{minipage}{0.44\columnwidth}
    \centering
    \footnotesize
    $\vec{c} = \vec{a} \ccorr \vec{b}$

    \begin{align*}
      c_0 & = a_0b_0 + a_1b_1 + a_2b_2\\
      c_1 & = a_0b_2 + a_1b_0 + a_2b_1\\
      c_2 & = a_0b_1 + a_1b_2 + a_2b_0
    \end{align*}
  \end{minipage}
  \caption{Circular correlation as compression of the tensor product. Arrows indicate
    summation patterns, nodes indicate elements in the tensor product. Adapted from~\protect\cite{plate1995holographic}.\label{fig:ccorr}}
\end{figure}

As a compositional operator, circular correlation can be interpreted as a
compression of the tensor product. While the tensor product assigns a separate
component $c_{ij} = a_ib_j$ for each pairwise interaction of entity features, in correlation each
component corresponds to a \emph{sum} over a fixed partition of pairwise
interactions (see also \cref{fig:ccorr}).
Intuitively, a feature in the tuple representation is ``on'' if at least one
partition of subject-object-interactions is on.
This form of compression can be very effective since it allows to share weights
in $\vec{r}_p$ for semantically similar interactions. For example, to model
relational patterns in the \textit{partyOf} relation, it might be sufficient to
know whether subject and object are a \textit{liberal person and liberal party}
OR if they are a \textit{conservative person and conservative party}. These
interactions can then be grouped in the same partition.
Additionally, it is typically the case that only a
subset of all \emph{possible} interactions of latent features are relevant to model
relational patterns. Irrelevant interactions can then be grouped in the same
partitions and collectively be assigned a small weight in $\vec{r}_p$.
                            %
                            %
Please note that the partitioning is not learned but fixed in advance through
the correlation operator. This is possible because the entity representations
are learned and the latent features can thus be ``assigned'' to the best
partition during learning.

Compared to the tensor product, circular correlation has the important advantage
that it does not increase the dimensionality of the composite representation
(see also \cref{fig:nn:hole}).
The memory complexity of the tuple representation is therefore linear in the
dimensionality $d$ of the entity representations. Moreover, the runtime complexity is 
quasilinear (loglinear) in $d$, as circular correlation can be computed via
\begin{align*}
  \va \ccorr \vb & = \ifft \left( \conj{\fft(\va)} \hadamard \fft(\vb) \right)
\end{align*}
where $\fft(\cdot)$ and $\ifft(\cdot)$ denote the \emph{fast Fourier transform} (FFT)
and its inverse, $\conj{\vx}$ denotes the complex conjugate of ${\vx \in \mathbb{C}^d}$,
and $\hadamard$ denotes the Hadamard (entrywise) product.
The computational complexity of the FFT is $\BigO(d \log d)$.
\Cref{tab:complexity} summarizes the improvements of circular correlation over the
tensor product. \Cref{tab:params} lists the memory complexity of \hole in comparison to
other embedding models.

Circular convolution $\cconv : \R^d \times \R^d \to \R^d$ is an operation that
is closely related to circular correlation and defined as
\begin{equation}
  [\vec{a} \cconv \vec{b}]_k = \sum_{i=0}^{d-1} a_i b_{(k - i) \operatorname{mod} d} .
  \label{eq:cconv}
\end{equation}
In comparison to convolution, correlation has two main advantages when used as a
compositional operator:
\begin{description}
\item[Non Commutative] Correlation, unlike convolution, is not commutative,
  i.e., $\va \ccorr \vb \neq \vb \ccorr \va$. Non-commutativity is necessary to
  model asymmetric relations (directed graphs) with compositional representations.
\item[Similiarity Component] In the correlation $\va \ccorr \vb$, a single
  component $[\va \ccorr \vb]_0 = \sum_i {a_i}{b_i}$ corresponds to the dot product $\langle
  \va, \vb \rangle$. The existence of such a component can be helpful to model
  relations in which the similarity of entities is important. No
  such component exists in the convolution $\va \cconv \vb$ (see also fig. 1 in the
  supplementary material).
\end{description}

\noindent
To compute the representations for entities and relations, we minimize either
\cref{eq:loss} or \eqref{eq:margin-loss} via stochastic gradient descent (SGD).
Let ${\theta \in \{\ve_i\}_{i=1}^{n_e} \cup \{\vr_k\}_{k=1}^{n_r}}$ denote the embedding 
of a single entity or relation and let ${f_{spo} = \sigma(\vr_p^\transp(\ve_s \ccorr
  \ve_o))}$. 
The gradients of \cref{eq:hole} are then given by
\begin{equation*}
  \frac{\partial f_{spo}}{\partial \theta} = \frac{\partial f_{spo}}{\partial
    \eta_{spo}} \frac{\partial \eta_{spo}}{\partial \theta},
\end{equation*}
where
\begin{align}
  \label{eq:hole:grad}
  \frac{\partial \eta_{spo}}{\partial \vr_p} = \ve_s \ccorr \ve_o, 
  &&
  \frac{\partial \eta_{spo}}{\partial \ve_s} = \vr_p \ccorr \ve_o, 
  &&
  \frac{\partial \eta_{spo}}{\partial \ve_o} = \vr_p \cconv \ve_s .
\end{align}
The partial gradients in \cref{eq:hole:grad} follow directly from
\begin{align}
  \label{eq:omni}
  \vr_p^\transp (\ve_s \ccorr \ve_o) = \ve_s^\transp(\vr_p \ccorr \ve_o) = \ve_o^\transp(\vr_p \cconv \ve_s)
\end{align}
and standard vector calculus. 
\Cref{eq:omni} can be derived as follows:
First we rewrite correlation in terms of convolution:
\begin{equation*}
  \label{eq:corr-conv}
  \va \ccorr \vb = \invl{\va} \cconv \vb
\end{equation*}
where $\invl{\va}$ denotes the \emph{involution} of $\va$, meaning that
$\invl{\va}$ is the mirror image of $\va$ such that $\invl{a}_i =
a_{-i \operatorname{mod} d}$~\citep[eq. 2.4]{schonemann1987some}.
\Cref{eq:omni} follows then from the following identities in convolution algebra~\citep{plate1995holographic}:
\begin{align*}
  \label{eq:corr-id}
  \vc^\transp(\invl{\va} \cconv \vb) = \va^\transp (\invl{\vc} \cconv \vb); 
  &&
  \vc^\transp(\invl{\va} \cconv \vb) = \vb^\transp (\va \cconv \vc) .
\end{align*}
Similar to correlation, the circular convolution in \cref{eq:hole:grad} can be
computed efficiently via ${\va \cconv \vb = \ifft(\fft(\va)
  \hadamard \fft(\vb))}$.

\section{Associative Memory}
\label{sec:assoc-mem}

In this section we outline the connections of \cref{eq:hole} and
\cref{eq:hole:grad} to holographic models of associative memory. Such models
employ a sequence of convolutions and correlations as used in holography to
store and retrieve information (e.g.,
see~\cite{gabor1969associative,poggio1973holographic}).
In particular, holographic reduced representations~\citep{plate1995holographic}
\emph{store} the association of $\va$ with $\vb$ via their circular convolution 
\[
  \vec{m}  = \va \conv \vb ,
\]
and \emph{retrieve} the item associated with $\va$ from $\vec{m}$ via
\[
  \vb^\prime \approx \va \corr \vec{m} = \vb \conv (\va \corr \va)
\]
If $\va \corr \va \approx \delta$ (the identity element of convolution),
it holds that $\vb \approx \vb^\prime$ and we can retrieve a noisy version of $\vb$. 
For denoising, we can pass the retrieved vector
through a clean-up memory, which returns the stored item with
the highest similarity to the item retrieved from
$\vec{m}$. For instance, if $\|\va\| = \|\vb_i\| = 1$, we can perform the clean-up via
\begin{equation}
  \label{eq:cleanup}
  \vb = \argmax_{\vb_i} \vb_i^\transp(\va \ccorr \vec{m})
\end{equation}
Multiple elements are stored in $\vec{m}$ via superposition:
\[
  \vec{m} = \sum\nolimits_i \va_i \conv \vb_i.
\]
Hence, $\vec{m}$ acts in this scheme as a memory that stores associations
between vectors which are stored and retrieved using circular convolution and
correlation.

Consider now the following model of associative memory for relational data:
Let $\Set{S}_o = \setdef{(s, p)}{\phi_p(\Ss, \So) = 1}$ be the set of all
subject-predicate indices for which the relation $\SR_p(\Ss, \So)$ is true. 
Next, we store these existing relations via convolution and superposition
in the representation $\ve_o$:
\begin{equation}
  \label{eq:rel-assoc}
  \ve_o = \sum_{(s,p) \in \Set{S}_o} \vr_p \cconv \ve_s 
\end{equation}
In this scheme, the compositional representation $\ve_s \ccorr \ve_o$ of \cref{eq:hole-comp}
would be analogous to retrieving the stored predicates $p$ that exist between $\Ss$ and
$\So$. Similarly, computing $\sigma(\vr_p^\transp(\ve_s \ccorr \ve_o))$ as in 
\cref{eq:hole} is analogous to computing the probability that $\vr_p$ 
is included in the retrieved relations, i.e., that we have seen the 
triple ${\SR_p(\Ss,\So)}$. The norm constraints of
\cref{eq:cleanup} can either be enforced directly (by projection the embeddings
onto the unit circle) or through the regularization of the embeddings (which is
equivalent to $\|\ve_i\| \leq C_e$, ${\|\vr_k\| \leq C_r}$, where $C_e,C_r$ depend
on the regularization parameter).

An important difference of \hole to associative memory is that 
it does not only memorize, but it generalizes in a well defined way:
In associative memory we are given the embeddings and store the associations
directly, typically via Hebbian learning (e.g., see \cref{eq:rel-assoc}). 
In \hole, we do not simply store the associations, but instead 
learn the embeddings that best explain the observed data. 
By iterating over $\Set{D}$ with SGD, we update the
embeddings of the objects via
\begin{equation}
  \label{eq:grad-update}
  \ve_o^{t+1} \gets \ve_o^t - \mu \frac{\partial \loss}{\partial f}\frac{\partial f}{\partial \eta}( \vr_p^t
  \cconv \ve_s^t),
\end{equation}
where $\mu$ denotes the learning rate.
Please note that \cref{eq:grad-update} is analogous to the association of
predicate and subject in holographic associative memory.
Hence, we can interpret \cref{eq:grad-update} as adapting the ``memory''
$\ve_o$, such that the retrieval of the observed facts is improved. The same
analogy holds for the updates of $\ve_s$ and $\vr_p$, however with the roles of
correlation and convolution in storage and retrieval reversed.
Moreover, in minimizing \cref{eq:loss} via \cref{eq:grad-update}, 
we are estimating a probability distribution over possible states of the
knowledge graph which allows us to predict the probability of any
\emph{possible} triple in the graph \cite{nickel2015review}.

\section{Experiments}

\subsection{Knowledge Graphs}
\label{sec:exp-lp}

\begin{table*}[bt]
  \centering
  \captionof{table}{Results for link prediction on WordNet (WN18), Freebase
    (FB15k) and Countries data.\label{tab:lp}}
  \small
  \subfloat[]{\label{tab:lp:kgs}
    \begin{tabular}{lllllllllll}
      \toprule
      & \multicolumn{5}{c}{\textbf{WN18}} & \multicolumn{5}{c}{\textbf{FB15k}} \\
      \cmidrule(r){2-6} \cmidrule(l){7-11}
      & \multicolumn{2}{c}{MRR} & \multicolumn{3}{c}{Hits at} & \multicolumn{2}{c}{MRR} & \multicolumn{3}{c}{Hits at} \\
      \cmidrule(r){2-3} \cmidrule(lr){4-6} \cmidrule(lr){7-8} \cmidrule(l){9-11}
      \textbf{Method} & Filter & Raw & 1 & 3 & 10 & Filter & Raw & 1 & 3 & 10 \\
      \midrule
      \transe & 0.495 & 0.351 & 11.3 & 88.8 & 94.3
                                                                                        & 0.463 & 0.222 & 29.7 & 57.8 & \textbf{74.9} \\
      \transr  & 0.605 & 0.427 & 33.5 & 87.6 & 94.0 
                                                                                        & 0.346 & 0.198 & 21.8 & 40.4 & 58.2 \\
      \emlp & 0.712 & 0.528 & 62.6 & 77.5 & 86.3 
                                                                                        & 0.288 & 0.155 & 17.3 & 31.7 & 50.1 \\
      \rescal  & 0.890 & 0.603 & 84.2 & 90.4 & 92.8 
                                                                                        & 0.354 & 0.189 & 23.5 & 40.9 & 58.7 \\
      \midrule
      \hole & \textbf{0.938} & \textbf{0.616} & \textbf{93.0} &\textbf{94.5} & \textbf{94.9}
                                                                                        & \textbf{0.524} & \textbf{0.232} & \textbf{40.2} & \textbf{61.3} & 73.9 \\
      \bottomrule
    \end{tabular} 
  }
  \hfill
  \subfloat[]{\label{tab:lp:countries}
    \begin{tabular}{llll}
      \toprule
      & \multicolumn{3}{c}{\textbf{Countries}} \\
      \cmidrule(l){2-4}
      & \multicolumn{3}{c}{AUC-PR} \\
      \cmidrule(l){2-4}
      \textbf{Method} & S1 & S2 & S3 \\
      \midrule
      Random & 0.323 & 0.323 & 0.323 \\
      Frequency & 0.323 & 0.323 & 0.308 \\
      \emlp & 0.960 & 0.734 & 0.652 \\
      \rescal & \textbf{0.997} & 0.745 & 0.650 \\
      \midrule
      \hole & \textbf{0.997} & \textbf{0.772} & \textbf{0.697} \\
      \bottomrule
    \end{tabular}
  }
  \vspace{-1em}
\end{table*}

To evaluate its performance for link prediction on knowledge graphs, 
we compared \hole to state-of-the-art models on two commonly used benchmark
datasets for this task:
\begin{description}
\item[WN18] WordNet is a KG that groups words into synonyms
  and provides lexical relationships between words. The WN18 dataset
  consists of a subset of WordNet, containing 40,943 entities, 18 relation
  types, and 151,442 triples.
\item[FB15k] Freebase is a large knowledge graph that stores general facts
  about the world (e.g., harvested from Wikipedia, MusicBrainz, etc.). The FB15k dataset consists of a subset of Freebase, containing
  14,951 entities, 1345 relation types, and 592,213 triples.
\end{description}
For both datasets we used the fixed training-, validation-, and test-splits
provided by \citet{bordes2013translating}. 
          %
As baseline methods, we used \rescal, \transe, \transr, and \emlp. 
To facilitate a fair comparison we reimplemented \emph{all} models and
used the identical loss and optimization method for training, i.e., 
SGD with AdaGrad~\citep{duchi2011adaptive} and the ranking loss of
\cref{eq:margin-loss}. This improved the results of \transe and
\rescal significantly on both datasets compared to results reported by \citet{bordes2013translating}.\footnote{\transe in its original implementation used SGD without
  AdaGrad. \rescal used the least-squares loss and ALS updates.}

Following \citet{bordes2013translating}, we generated negative relation
instances for training by corrupting positive triples and used the
following evaluation protocol: For each true triple $\SR_p(\Ss, \So)$
in the test set, we replace the subject $\Ss$ with each entity $\Ss^\prime \in
\SE$, compute the score for $\SR_p(\Ss^\prime, \So)$, and rank all these instances
by their scores in decreasing order. Since there can exist multiple true
triples in the test set for a given predicate-object pair, we remove
all instances from the ranking where $\SR_p(\Ss^\prime, \So) = 1$ and ${\Ss \neq
  \Ss^\prime}$, i.e., we consider only the ranking of the test
instance among all wrong instances (which corresponds to the ``Filtered''
setting in \citet{bordes2013translating}). We then repeat this procedure by
replacing the object $\So$. To measure the quality of the ranking, we use the
mean reciprocal rank (MRR) which is commonly used in information retrieval
and in contrast to mean rank is less sensitive to outliers.
In addition to MRR, we report the ratio in which $\SR_p(\Ss, \So)$
occurs within the first $n$ results (denoted by ``Hits at $n$'').
We optimized the hyperparameters of all models via extensive grid search and
selected the model with the best filtered MRR score on the validation set.
The results of these experiments are shown in \cref{tab:lp:kgs}.
It can be seen that \hole is able to outperform the considered baseline methods
significantly and consistently on both datasets. For instance, \transe and
\transr rank the test instance only in 11.5\% and 33.5\% of the cases as the
most likely triple in WN18 (\mbox{Hits at 1}). In contrast, \hole ranks the test
instance in 93.0\% of the cases as the most likely instance. While less
pronounced, similar results can be observed on FB15k.
In \cref{tab:params}, we report the dimensionality $d$ and the resulting
number of parameters of the selected models. It can be seen that \hole is far
more efficient in the number of parameters compared to the tensor product model
\rescal. Although the dimensionality $d$ of the \hole embedding is larger than
\textsc{Rescal}'s (what is to be expected due to the compressive effect of
correlation), the overall number of parameters is significantly reduced as its
memory complexity depends only linearly on $d$.
Also, \hole is typically very fast to compute. On standard hardware (Intel
Core(TM) i7U 2.1GHz) and for $d=150$ (as used in the experiments) the runtime to
compute the probability of a single triple is around 40$ \mu$s. To compute all
embeddings, a single epoch on WN18 takes around 11s (earlier epochs are slower
since more examples violate the margin). Typically, we need 200-500 epochs
(depending on the dataset) to arrive at the best estimates for the embeddings.

\subsection{Relational Learning}
We have shown that \hole can predict triples successfully in knowledge
graphs. In additional experiments, we wanted to test the relational learning
capabilities of the compositional representation. For this purpose, we used the
countries dataset of \citet{bouchard2015approximate}, which consists of 244
countries, 22 subregions (e.g., \textit{Southern Africa}, \textit{Western
  Europe}) and 5 regions (e.g., \textit{Africa}, \textit{Americas}). 
Each country is located in exactly one region and subregion, each subregion is
located in exactly one region, and each country can have a number of other
countries as neighbors. From the raw data we created a relational representation
with two predicates: \textit{locatedIn($e_1$, $e_2$)} and
\mbox{\textit{neighborOf($e_1$, $e_2$)}}.
The task in the experiment was to predict $\textit{locatedIn}(c, r)$ instances,
where $c$ ranges over all countries and $r$ over all regions in the data.
The evaluation protocol was the following: First, we split all countries randomly in
train (80\%), validation (10\%), and test (10\%) set, such that for each country
in the test set there is at least one neighbor in the training set.
Next, we removed triples from the test and validation set
in three different settings:
\begin{description}
\item[S1)] In the basic setting we only set $\textit{locatedIn}(c,r)$ to missing for
  countries in the test/valid. set. In this
  setting, the correct relations can be predicted from patterns of the form:
  \[
    \textit{locatedIn}(c, s) \land \textit{locatedIn}(s, r) \lif \textit{locatedIn}(c, r)
  \]
  where $s$ refers to the country's subregion.
\item[S2)] In addition to the triples of S1, we set
  $\textit{locatedIn}(c, s)$ to missing for all
  countries $c$ in the test/valid. set and all subregions $s$ in the data. 
  In this setting, the correct triples can be predicted from:
  \[
    \textit{neighborOf}(c_1, c_2) \land \textit{locatedIn}(c_2, r) \lif \textit{locatedIn}(c_1, r)
  \]
  This is a harder task than S1, since a country can have multiple
  neighbors and these can be in different regions.
\item[S3)] In addition to the triples of S1 and S2 we set
  $\textit{locatedIn}(n, r)$ to missing for all neighbors $n$ of all countries in the test/valid. set
  and all regions $r$ in the data. In this setting, the correct
  triples can be predicted from: 
  \begin{multline*}
    \textit{neighborOf}(c_1, c_2) \land \textit{locatedIn}(c_2, s)\, \land \\
    \textit{locatedIn}(s, r) \lif \textit{locatedIn}(c_1, r)
  \end{multline*}
  This is the most difficult task, as it not only involves the
  \textit{neighborOf} relation, but also a path of length 3.
\end{description}
See \cref{fig:countries} for an illustration of the data structure and the test
settings. 
We measured the prediction quality via the area under the
precision-recall curve (AUC-PR).
The results of the experiments are shown in \cref{tab:lp:countries}. It
can be seen that \hole is very successful in these
learning tasks. For S1, the missing triples are predicted nearly
perfectly. Moreover, even for the most difficult task S3, \hole achieves very good
results, especially since not every country's region can be predicted from its
neighbors (e.g., islands have no neighbors). The poorer results of \rescal and
\emlp can likely be explained with overfitting (although the models are
regularized), since the difference to \hole is
reduced when the hyperparameters are optimized on the
test set instead of the validation set.
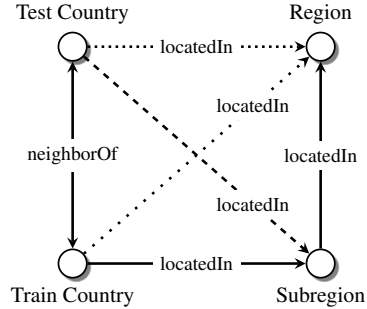
\begin{figure}[tb]
  \centering
  \vspace{-1ex}
  \resizebox{0.6\columnwidth}{!}{
    \begin{tikzpicture}[scale=2.0,baseline,thick,node distance=.5cm]
      \GraphInit[vstyle=Classic]
      \tikzset{vertex/.style =
        {draw=black,shape=circle,fill=white,minimum size=13pt,circular
          drop shadow}}
      \tikzset{edge/.style = {->,>=stealth,very thick}}
      \tikzset{neigh/.style = {<->,>=stealth,very thick}}
      \node[vertex] at (0,1.75)[label=above:Region] (r) {};
      \node[vertex] at (0,0)[label=below:Subregion] (sr) {};
      \node[vertex] at (-2,1.75)[label=above:Test Country] (ts) {};
      \node[vertex] at (-2,0)[label=below:Train Country] (tr) {};
    
      \draw[edge,dashed] (ts) -- (sr) node [near end, fill=white] {\footnotesize{locatedIn}};
      \draw[edge,dotted] (ts) -- (r) node [midway, fill=white] {\footnotesize{locatedIn}};
      \draw[edge,loosely dotted] (tr) -- (r) node [near end, fill=white] {\footnotesize{locatedIn}};
      \draw[edge] (tr) -- (sr) node [midway, fill=white] {\footnotesize{locatedIn}};
      \draw[neigh] (tr) -- (ts) node [midway, fill=white] {\footnotesize{neighborOf}};
      \draw[edge] (sr) -- (r) node [midway, fill=white] {\footnotesize{locatedIn}};
    \end{tikzpicture}
  }
  \hspace{1.5em}
  \caption{Removed edges in countries experiment: S1) dotted
    S2) dotted and dashed S3) dotted, dashed and loosely dotted.\label{fig:countries}}
\end{figure}
We observed similar results as in this experiment on commonly used benchmark datasets for
statistical relational learning. Due to space constraints, we report these
experiments in the supplementary material.

\section{Conclusion and Future Work}
In this work we proposed \hole, a compositional vector space model
for knowledge graphs that is based on the circular correlation of vectors.
An attractive property of circular correlation in this context is that it
creates fixed-width representations, meaning that the compositional
representation has the same dimensionality as the representation of its
constituents. 
In \hole, we exploited this property to create a compositional
model that can capture rich interactions in relational data but simultaneously
remains efficient to compute, easy to train, and very scalable. 
Experimentally we showed that \hole provides state-of-the-art performance on a
variety of benchmark datasets and that it can model complex relational patterns
while being very economical in the number of its parameters.
Moreover, we highlighted connections of \hole to holographic models of
associative memory and discussed how it can be interpreted in this context. This
creates not only a link between relational learning and associative memory, but
also allows for principled ways to query the model, for instance in question
answering.
In future work we plan to further exploit the fixed-width representations of
holographic embeddings in complex scenarios, since they are
especially suitable to model higher-arity relations (e.g., \textit{taughtAt(John, AI,
  MIT)}) and facts about facts (e.g., \mbox{\textit{believes(John, loves(Tom,
    Mary))}}).

\subsection*{Acknowledgments \& Reproducibility}
This material is based upon work supported by the Center for Brains, Minds and
Machines (CBMM), funded by NSF STC award CCF-1231216. 
The code for models and experiments used in this paper is available at
\url{https://github.com/mnick/holographic-embeddings}.

{
  \bibliographystyle{aaai}
  \bibliography{hole}
}

\end{document}